\documentclass[sigconf]{acmart}
\usepackage{kotex}
\usepackage{adjustbox}
\usepackage{subfigure}
\usepackage{framed,multirow}
\usepackage{algorithm}
\usepackage{algpseudocode}
\usepackage{xcolor}
\usepackage{soul} 
\usepackage{threeparttable} 

\usepackage{caption} 
\usepackage{stfloats} 

\AtBeginDocument{%
  \providecommand\BibTeX{{%
    \normalfont B\kern-0.5em{\scshape i\kern-0.25em b}\kern-0.8em\TeX}}}
    
\newcommand\HH{
  \global\let\savedtextbullet\textbullet
  \gdef\textbullet{%
    \par\noindent\savedtextbullet\global\let\textbullet\savedtextbullet
  }%
}

\copyrightyear{2022}
\acmYear{2022}
\setcopyright{acmcopyright}
\acmConference[WWW '22] {Proceedings of the ACM Web Conference 2022}{April 25--29, 2022}{Lyon, France.}
\acmBooktitle{Proceedings of the ACM Web Conference 2022 (WWW '22), April 25--29, 2022, Lyon, France}
\acmPrice{15.00}
\acmISBN{978-1-4503-9096-5/22/04}
\acmDOI{10.1145/3485447.3511978}

\begin{document}

\title{Semi-Siamese Bi-encoder Neural Ranking Model Using Lightweight Fine-Tuning}

\author{Euna Jung}
\authornote{Both authors contributed equally to this research.}
\affiliation{%
	\institution{GSCST \\ 
	Seoul National University}
	\streetaddress{}
	\city{Seoul}
	\country{Korea}
}
\email{xlpczv@snu.ac.kr}

\author{Jaekeol Choi}
\authornotemark[1]
\affiliation{%
	\institution{Seoul National University \\ \& Naver Corp.}
	\city{Seoul}
	\country{Korea}
}
\email{jaekeol.choi@snu.ac.kr}


\author{Wonjong Rhee}
\affiliation{%
	\institution{GSCST, GSAI, AIIS \\
	Seoul National University}
	\city{Seoul}
	\country{Korea}
}
\email{wrhee@snu.ac.kr}

\renewcommand{\shortauthors}{Euna Jung, Jaekeol Choi, \& Wonjong Rhee} 

\begin{abstract}

A BERT-based Neural Ranking Model (NRM) can be either a cross-encoder or a bi-encoder. Between the two, bi-encoder is highly efficient because all the documents can be pre-processed before the actual query time. 
In this work, we show two approaches for improving the performance of BERT-based bi-encoders. The first approach is to replace the full fine-tuning step with a lightweight fine-tuning. We examine lightweight fine-tuning methods that are adapter-based, prompt-based, and hybrid of the two. The second approach is to develop semi-Siamese models where queries and documents are handled with a limited amount of difference. 
The limited difference is realized by learning two lightweight fine-tuning modules, where the main language model of BERT is kept common for both query and document.  
We provide extensive experiment results for monoBERT, TwinBERT, and ColBERT where three performance metrics are evaluated over Robust04, ClueWeb09b, and MS-MARCO datasets. The results confirm that both lightweight fine-tuning and semi-Siamese are considerably helpful for improving BERT-based bi-encoders. In fact, lightweight fine-tuning is helpful for cross-encoder, too.\footnote{The code is available at \url{https://github.com/xlpczv/Semi_Siamese}}

\end{abstract}



\maketitle
\section{Introduction}

Since the advent of large-scale language models, BERT-based Neural Ranking Models (NRMs) \cite{nogueira2019passage, khattab2020colbert, lu2020twinbert} have been developed and shown to achieve state-of-the-art performance. A BERT-based NRM can be classified either as a cross-encoder or as a bi-encoder. Although cross-encoders generally outperform bi-encoders, bi-encoders are superior in terms of computational efficiency because they allow one-time pre-processing of the long documents. Therefore, bi-encoders tend to receive more attention from industrial practitioners.
For processing both queries and documents, a bi-encoder uses a common BERT model with a fixed set of weight values. This has been considered to be a mandatory requirement, because the underlying language model is desired to be the same for handling both queries and documents and because heterogeneous models indeed show very poor performance. Therefore, all of the existing bi-encoder models are \textit{Siamese models}.

\begin{table}[t]
\caption{Query and document lengths - average and standard deviation of word counts are shown for three IR datasets.}
\label{tab:query_length}
\small
\adjustbox{max width=\linewidth}{%
\begin{tabular}{c|c|c}
    \toprule
    \textbf{Dataset} &  \textbf{Query word count} & \textbf{Document word count} \\
    \midrule
    Robust04 & 2.66 ($\pm$0.69) & 912 ($\pm$2114)    \\ 
    ClueWeb09b & 2.40 ($\pm$0.98) & 2346 ($\pm$2128) \\ 
    MS-MARCO & 6.00 ($\pm$2.58) & 2454 ($\pm$4761)   \\
    \bottomrule
\end{tabular}}
\end{table}

\begin{table}[t]
\caption{Query examples - Robust04 and ClueWeb09b mainly contain short and keyword-based queries, but MS-MARCO mainly contains long and descriptive queries.}
\label{tab:query_examples}
\small
\adjustbox{max width=\linewidth}{%
\begin{tabular}{c|c|c}
    \toprule
    \textbf{Dataset} & \multicolumn{2}{c}{\bf{Query examples}} \\
    \midrule
    \multirow{2}{*}{Robust04} & \multicolumn{2}{l}{new fuel sources} \\
             & \multicolumn{2}{l}{most dangerous vehicles} \\
    \midrule
    \multirow{2}{*}{ClueWeb09b} & \multicolumn{2}{l}{dinosaurs} \\
               & \multicolumn{2}{l}{air travel} \\
    \midrule
    \multirow{2}{*}{MS-MARCO} & \multicolumn{2}{l}{Vitamin D deficiency and skin lesions} \\
    & \multicolumn{2}{l}{What happens when blood goes through the lungs?} \\
    \bottomrule
\end{tabular}}
\end{table}

Learning a ranking model is a special task because of the involvement of query and document. In particular, query and document can have distinct characteristics.   
Table~\ref{tab:query_length} summarizes the length information of query and document for three popular Information Retrieval (IR) datasets. It can be immediately noticed that the length difference between query and document is remarkably large. Robust04 and ClueWeb09b contain very short queries while MS-MARCO has relatively longer queries. Even though not shown in the table, many of ClueWeb09b queries consist of only one word per query. On the other hand, a document usually contains more than 1,000 words for all three datasets. Table~\ref{tab:query_examples} shows examples of the queries. It can be seen that Robust04 and ClueWeb09b have keyword-based queries while MS-MARCO has descriptive queries in full or almost full sentence format. Because the documents of the three datasets are usually in full sentence format, it can be concluded that Robust04 and ClueWeb09b have different sentence formats for query and document while MS-MARCO has the same format. In this work, we hypothesize that a high-performance bi-encoder should process query and document with two different networks, because they tend to have distinct characteristics.

Because heterogeneous networks do not perform well\footnote{
We confirmed this by fine-tuning query and document models independently. For example, heterogeneous full fine-tuning of ColBERT on Robust04 resulted in 0.3233~(P@20) while Siamese full fine-tuning resulted in 0.3355~(P@20) where the p-value was 0.014.
}, we propose Semi-Siamese~(SS) bi-encoder neural ranking models that can properly reflect the different characteristics of query and document. Our semi-Siamese networks are based on a common pre-trained BERT model that is not fine-tuned at all. Instead, a mild differentiation between the query network and the document network is implemented through a lightweight fine-tuning method including prompt-tuning~\cite{lester2021power}, prefix-tuning~\cite{li2021prefix}, and LoRA~\cite{hu2021lora}. The resulting semi-Siamese networks have less than 1\% difference in terms of the number of parameters that are different. We also introduce LoRA$+$ that allows a small additional differentiation and also consider two \textit{sequential hybrids} of prefix-tuning and LoRA.   


While semi-Siamese learning for bi-encoders is our main focus, we also investigate the benefits of lightweight fine-tuning for Siamese cross-encoders and Siamese bi-encoders. With our best knowledge, we are the first to apply lightweight fine-tuning for improving NRMs. Originally, lightweight fine-tuning methods were developed to reduce task-specific parameter memory/storage and computational cost. But, we will show that they can also provide performance enhancement through their regularizing effect of NRMs. Compared to a full fine-tuning, a lightweight fine-tuning of NRM allows only a limited amount of parameters to be modified and we obtain performance improvements by choosing adequate lightweight fine-tuning methods. Our method improves bi-encoders that are practical in the real web search environment. 

Our contributions can be summarized as below.
\begin{itemize}
    \item For cross-encoder, we show that adapter-based lightweight fine-tuning methods (LoRA and LoRA$+$) can improve the performance by 0.85\%-5.29\%. 
    \item For bi-encoder, we show that prefix-tuning performs well for Robust04 and ClueWeb09b that have short queries. The improvement can be very large for TwinBERT, and a modest gain of 0.12\%-3.90\% is achieved for ColBERT.  
    \item For bi-encoder, we show that semi-Siamese learning is effective where the overall gain of 1.46\%-16.23\% is achieved for ColBERT.  
\end{itemize}

\section{Related Works}

\subsection{BERT-based NRMs}
With the advent of large-scale language models such as BERT and GPT, fine-tuning of such a pre-trained language model has become a standard approach for handling document ranking tasks. For neural ranking models, a variety of BERT-based NRMs have been developed. For instance, monoBERT \cite{nogueira2019passage} is often considered to be a powerful baseline that takes a pair of query and document as the input to the BERT model. 
Such cross-encoder models, however, are computationally demanding because the BERT's output representation needs to be calculated for every combination of query and document. If we have $N_q$ queries to evaluate for $N_d$ documents, this means BERT representations need to be evaluated $N_q \cdot N_d$ times. In contrast, bi-encoder models are computationally efficient because their BERT models do not jointly process each pair of query and document. With a bi-encoder, all of the $N_d$ documents are pre-processed only once, and their BERT representations are pre-stored. For each query, the query's BERT representation is calculated and used together with each of the pre-stored $N_d$ document representations for the relevance score calculation. Because the documents are much longer than the queries (see Table~\ref{tab:query_length}), the pre-processing of documents makes bi-encoders extremely efficient when compared to the cross-encoders. TwinBERT~\cite{lu2020twinbert} and ColBERT~\cite{khattab2020colbert} are two of the popular bi-encoder models. TwinBERT aggregates \textit{CLS} vectors of the query and the document for the relevance score estimation. ColBERT utilizes the interaction between the BERT representations instead. 


\subsection{Lightweight Fine-Tuning~(LFT)}
Traditionally, fine-tuning refers to updating all of the pre-trained weights of a neural network. Because all the weights are updated, we refer to it as \textit{full Fine-Tuning}~(FT) in this work. A full FT of BERT is not always necessary for achieving a high performance. Instead, \textit{partial fine-tuning} methods can be adopted for reducing task-specific parameter storage and computational cost. \citet{lee2019would} investigated a partial fine-tuning strategy where only the final layers are fine-tuned and showed that only a fourth of the final layers need to be fine-tuned to achieve 90\% of the original downstream task quality. \citet{radiya2020fine} showed that it sufficed to fine-tune only the most critical layers. Similar results can be found in \cite{liu2021autofreeze,zaken2021bitfit}. 
While a partial FT can be advantageous over a full FT, they both require the pre-trained BERT to be modified. This is not desirable especially when multiple downstream tasks need to be handled. Recently, another type of fine-tuning called \textit{Lightweight Fine-Tuning}~(LFT) has emerged. With an LFT, all of the BERT parameters are kept frozen. To implement the effect of fine-tuning, LFT instead augments the BERT model with small trainable elements. Two types of LFT have been shown to be extremely useful. In our work, the first type is addressed as \textit{adapter-based LFT} and it augments a language model like BERT with small weight modules. The second type is addressed as \textit{prompt-based LFT} and it augments a language model's representation vectors with a small number of input embedding vectors or mid-layer activation vectors. With our best knowledge, LFT has never been applied to neural ranking models before and we are the first to demonstrate its effectiveness for NRMs.


\subsubsection{Adapter-based LFT}

\citet{houlsby2019parameter} proposed an alternative of full FT where task-specific adapters were inserted between the layers of the pre-trained language model. The adapter's effectiveness was demonstrated over 26 diverse text classification tasks, where near full fine-tuning performance was achieved with only 3.6\% of additional parameters per task. Later, \citet{he2021effectiveness} showed that the adapter FT could mitigate forgetting issues of full FT because of the smaller weight deviations. LoRA~\cite{hu2021lora} is another example of adapter-based LFT where augmentation is implemented with a different design of small weight modules. Instead of inserting layers, LoRA expands the query and value projection weight matrices with linear low-rank residual matrices. Because of its simplicity and small size, we investigate only LoRA among the adapter-based LFTs.


\subsubsection{Prompt-based LFT}
GPT-3~\cite{brown2020language} is an enormously large model, and it is known for its capability of performing few-shot or zero-shot learning merely with text prompts. A survey of prompt-based learning can be found in \cite{liu2021pre}. Following the idea of prompt-based learning, \citet{lester2021power} introduced prompt-tuning where soft prompts were learned through back-propagation. Soft prompts are fine-tuned embeddings, and they do not correspond to discrete text embeddings. 
\citet{li2021prefix} extended the concept beyond the input layer to introduce prefix-tuning, where it was different from prompt-tuning in that it learned prefix activation vectors for all the layers including the input layer. In our work, we investigate both prompt-tuning and prefix-tuning for enhancing neural ranking models.

Adapter-based LFTs are fundamentally different from prompt-based LFTs. 
Adapter-based LFTs do not augment the representations at all, and prompt-based LFTs do not augment weights at all. This difference has a strong implication to NRMs as we will show in the experiment section. Despite of the difference, both are fine-tuning methods because their augmentations are task-specific, i.e. not dependent on input examples, where the trainable augmentation elements are fine-tuned for a specific downstream task.






\subsection{Semi-Siamese~(SS) Models}
All of the existing bi-encoder NRMs are Siamese models where a single version of fine-tuned BERT is used for processing both of the queries and the documents. Learning a heterogeneous bi-encoder model is also possible where two different versions of fine-tuned BERT models can be learned, one for processing queries and the other for processing documents. Such a heterogeneous bi-encoder, however, is known to suffer from a large performance degradation due to the deviation of the two language models for handling queries and documents. But when queries are short or when queries have significantly different characteristics compared to the documents, it makes sense to allow a certain degree of deviation to handle the queries better. A possible solution for learning is to introduce Semi-Siamese~(SS) models. Semi-Siamese models have never been used for information retrieval tasks, but they have been already adopted in image, video, and recommendation domains. \citet{du2020semi} used a semi-Siamese model to prevent over-fitting in a face recognition task with a small number of data examples. The model has two structurally identical networks that are trained simultaneously using different inputs.   
\citet{zhang2017iminet} proposed a semi-Siamese CNN network for vocal imitation search. To encode vocal imitation and real sound, two CNN networks are needed and they share lower layers. 
\citet{li2019semi} used a semi-Siamese model to make directional recommendations. First, two identical networks are trained with undirected data, and then the two networks are trained differently with directed data which makes them semi-Siamese. In the previous works, a semi-Siamese network is utilized when there is a need to train two slightly different networks. In our case of bi-encoder NRMs, we focus on the different characteristics between query and document and design semi-Siamese models to enhance the performance.

\section{Methodology}

\begin{figure*}[t]
    \centering
    \hfill
    \subfigure[Prefix-tuning]{\includegraphics[width=0.40\linewidth]{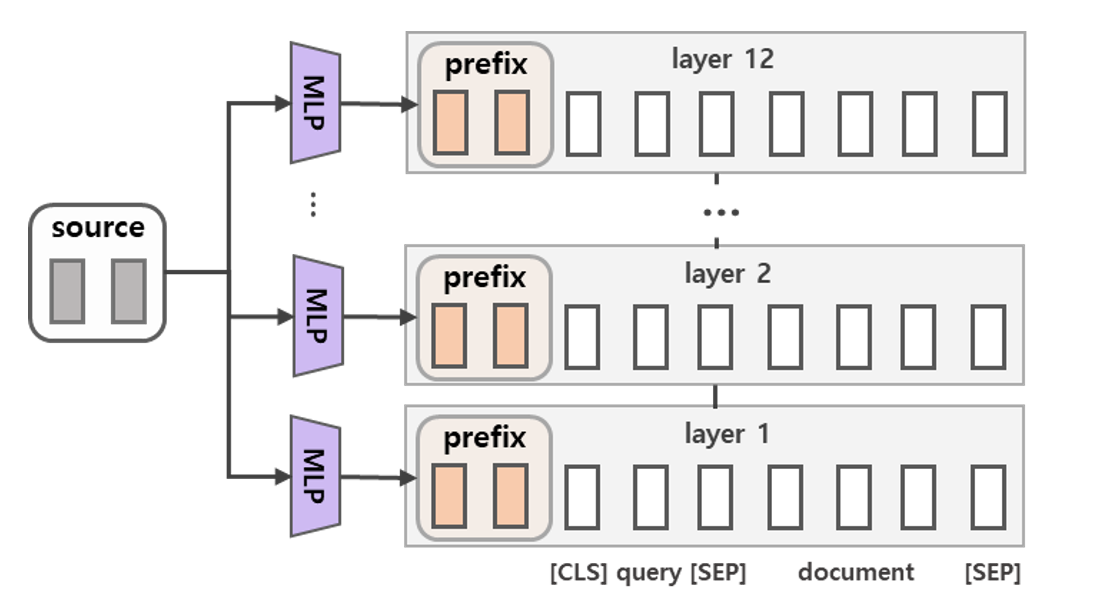}}
    \hfill
    \subfigure[LoRA]{\includegraphics[width=0.40\linewidth]{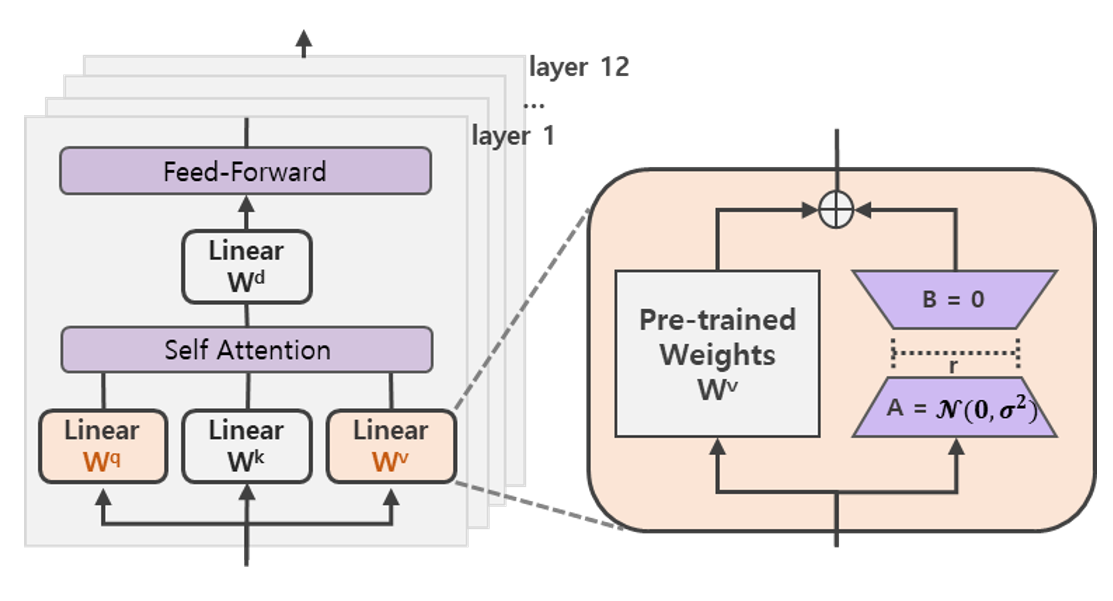}}
    \hfill
    \caption{Prefix-tuning and LoRA. (a) Prefix-tuning adds prefix embeddings to each layer. Each embedding is generated from the same source through layer-dependent MLPs. (b) LoRA affects $W^{q}$ and $W^{v}$ among the three weights involved in the self-attention module where $W^{q} = \{A^{q}, B^{q}\}$ and $W^{v} = \{A^{v}, B^{v}\}$ are LoRA weights for query and value, respectively. In addition to the two weights, LoRA$+$ also affects $W^{d}$ where $W^{d} = \{A^{d}, B^{d}\}$ are LoRA weights for the dense layer.
    }
    \label{fig:tuning_method}
\end{figure*}

\subsection{Document Re-ranking}
Document re-ranking is to rank a set of pre-selected documents for a given query according to the relevance score estimates, where a relevance score is estimated for each pair of query and document. Let $Q$ be a query consisting of tokens ${q_{1},q_{2}, ...,q_{|Q|}}$ and let $D$ be a document consisting of tokens ${d_{1},d_{2}, ...,d_{|D|}}$.
A positive pair of a query and a document $X_{pos}$ is comprised of relevant $Q$ and $D$ whereas a negative pair $X_{neg}$ is comprised of irrelevant $Q$ and $D$.
BERT-based NRMs are composed of two parts, BERT and the ranker. BERT processes query and document pairs $X$ to output contextualized representation vectors $Z$.
The ranker is a function $f: Z \mapsto s \in \mathbb{R}$ that estimates the relevance score $s$ using the BERT output.
$s_{pos}$ and $s_{neg}$ indicate the relevance scores of positive pair and negative pair, respectively.
NRMs are trained by minimizing the hinge loss of triplet data: 
\begin{equation}
    J(W) = \mathbb{E} (1 - \frac{e^{s_{pos}}}{e^{s_{pos}} + e^{s_{neg}}})
\end{equation}
To use the knowledge learned from pre-training tasks, we initialize BERT with the pre-trained weights and randomly initialize the ranker. During the training, we perform gradient updates on the trainable parameters for reducing the loss above. When performing a full fine-tuning, we update the entire set of BERT and ranker weights at the same time.

\subsection{Lightweight Fine-Tuning (LFT)}
Unlike the full fine-tuning that trains all of the BERT weights for a down-stream task, a lightweight fine-tuning trains only the augmented elements. While training only a small portion of parameters (1\% or less of BERT weights), some of the LFT methods have been proven to perform well, especially for NLG (Natural Language Generation) tasks. In this section, we address prefix-tuning and LoRA and explain how we apply these LFT methods to BERT-based NRMs.



\subsubsection{Prefix-tuning}
Prefix-tuning \cite{li2021prefix} trains the network to generate prefix activation vectors that are prepended to the normal activation vectors of the transformers.
Different from prompt-tuning that prepends prompts in front of the input word embeddings, prefix-tuning inserts prefixes to all layers including the input layer.
Where to prepend the prefix is a design choice that can affect the performance. {\citet{li2021prefix}} prepended the prefix to the key and value representations, but we chose to prepend the prefix to the representations right before the self-attention projection because we experimented with both options and found that our modified approach provided a better performance.
As shown in Figure~\ref{fig:tuning_method}(a), prefix embedding vectors $P_{\theta}$ that are generated from the source $P'_{\theta}$ are prepended to the normal activations of the layers.
The activation of an $i^{th}$ token $h_{i}$ in each layer becomes $P_{\theta}[i,:]$ if $i\in P_{idx}$ where $P_{idx}$ denotes the sequence of prefix indices and $P_{\theta}[i,:]$ is computed as the MLP output of the source vectors $P'_{\theta}[i,:]$ in low dimension (for simplicity, we omit layer indexes): 
\begin{equation}
    h_{i}=P_{\theta}[i,:]=MLP_{\theta}(P'_{\theta}[i,:]) \; \text{if} \; i \in P_{idx}
\end{equation}
We train both of the source $P'_{\theta}$ and the parameters of $MLP_{\theta}$ during the prefix-tuning. We set the length of the prefix as 10 and the source dimension as 768. $MLP_{\theta}$ consists of two linear layers between which a ReLU layer exists. The first linear layer down-projects the source vector corresponding to each index $P'_{\theta}[i,:]$ into the space of 256 dimensions and the second linear layer up-projects the vectors back into the space of 768 dimensions. The inference overhead induced by the prefix is less than 0.5\%.

When applying prefix-tuning on Siamese NRMs, we prepend the same prefixes to both models of query and document:
\begin{equation}
    h_{q,i}=h_{d,i}=P_{\theta}[i,:]
\end{equation}
For semi-Siamese NRMs, $h_{q,i}$ and $h_{d,i}$ are not constrained to be the same.




\subsubsection{LoRA}
LoRA stands for Low-Rank Adaptation~\cite{hu2021lora}, and it is an LFT method that freezes the pre-trained weights $W_{0}$ and trains only the rank decomposition matrix part of $\triangle W=BA$. As shown in Figure~\ref{fig:tuning_method}(b), the representation $h$ is computed as $h=(W_{0}+\triangle W)x=(W_{0}+BA)x$ where $x$ is the previous layer's representation vector. $W_{0}\in \mathbb{R}^{d\times k}$, $B\in \mathbb{R}^{d\times r}$, and $A\in \mathbb{R}^{r\times k}$ are the weight matrices, and $r\ll min(d,k)$. Because $r$ is much smaller than $d$ and $k$, the number of learnable parameters is significantly reduced from $d\times k$ to $r(d+k)$ for each projection process. Because LoRA is applied only to query and value matrices, the number of learnable parameters is decreased even further. Besides, we can sum up the additional LoRA weights to the original weights upon the completion of the training, and therefore the inference overhead can be forced to be zero.

\subsubsection{LoRA$+$}
NRM is a complicated task, and it can be desirable to increase the number of trainable parameters. After investigating several options, we have designed LoRA$+$ that is the same as LoRA except for additionally applying rank decomposition matrix to the dense layer $W^{d}$. Dense layer here refers to the layer following the self-attention layer in transformers. LoRA$+$ allows the model an additional room for fine-tuning.


\subsubsection{Sequential hybrid}
We additionally propose a new LFT method that combines prefix-tuning and LoRA. We devised our method based on the hypothesis that prefix-tuning and LoRA can complement each other. Note that they are involved in different parts of BERT. Prefix-tuning inserts task-specific information to the model by prepending activation vectors. On the other hand, LoRA fine-tunes the model by modifying projection weights through residual connections. 
There are many possibilities for combining the two, but we have chosen to sequentially combine them such that their learning dynamics are not mixed. After fine-tuning with one of the two LFT modules for $m$ epochs, we freeze the module. Then, we train the other LFT module for $n$ epochs. For robust04 and ClueWeb09b, we used $m=30$ and  $n=10$. For MS-MARCO, we used $m=10$ and $n=3$. Depending on which of prefix-tuning and LoRA is first fine-tuned, we end up with two different sequential hybrid LFTs. 
We present the \textit{Prefix-tuning $\rightarrow$ LoRA} in Algorithm~\ref{alg:ss_lora_prefix}, and \textit{LoRA $\rightarrow$ Prefix-tuning} is the same except for the order.

\begin{algorithm}
\caption{Sequential Hybrid: Prefix-tuning $\rightarrow$ LoRA}
\label{alg:ss_lora_prefix}
\begin{flushleft}
1. Train the prefix-tuning parameters for $m$ epochs and save the prefixes for the epoch with the best validation performance. \\
2. Freeze the prefixes with the saved prefixes. \\
3. Train the LoRA parameters for $n$ epochs and save the LoRA weights for the epoch with the best validation performance. \\
4. Freeze the LoRA weights with the saved LoRA weights. \\
\end{flushleft}
\end{algorithm}

\subsection{Semi-Siamese Neural Ranking Model}

\begin{figure*}[t]
    \centering
    \hfill
    \subfigure[SS Prefix-tuning]{\includegraphics[width=0.40\linewidth]{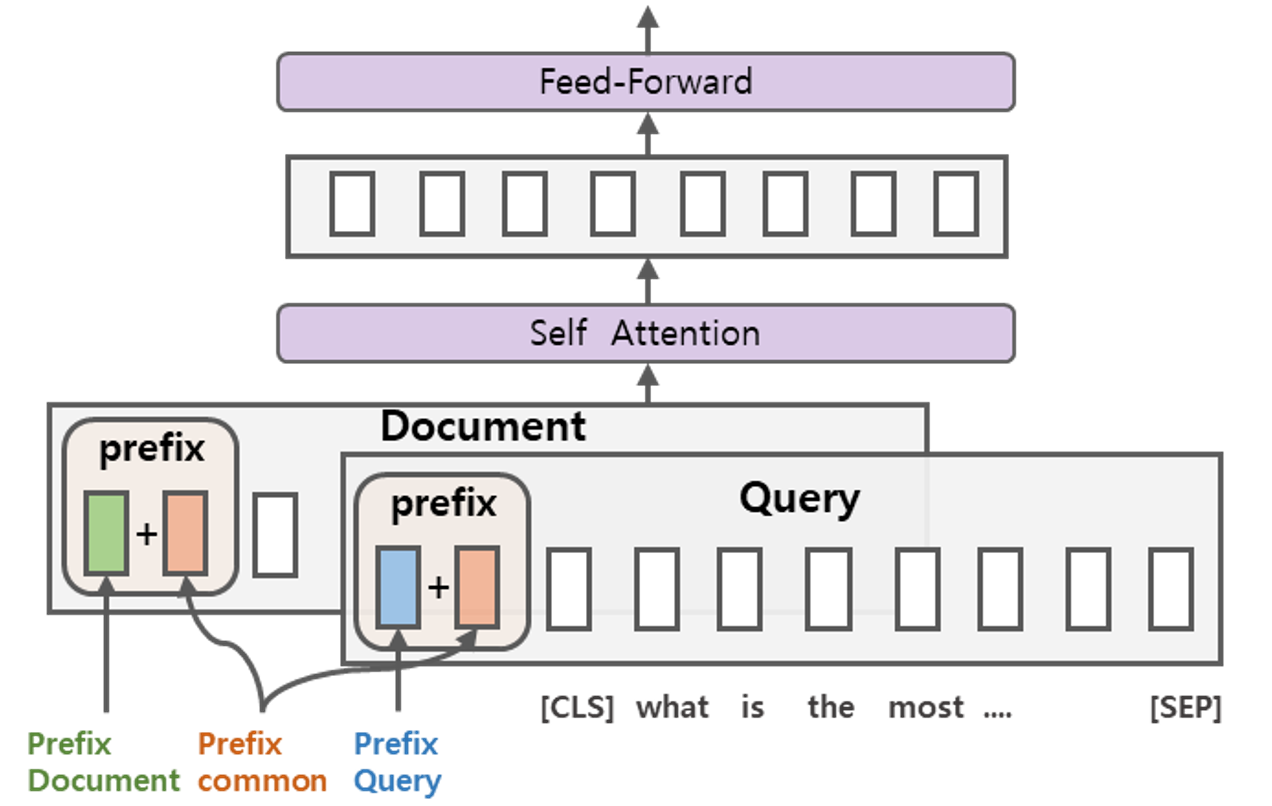}}
    \hfill
    \subfigure[SS LoRA]{\includegraphics[width=0.40\linewidth]{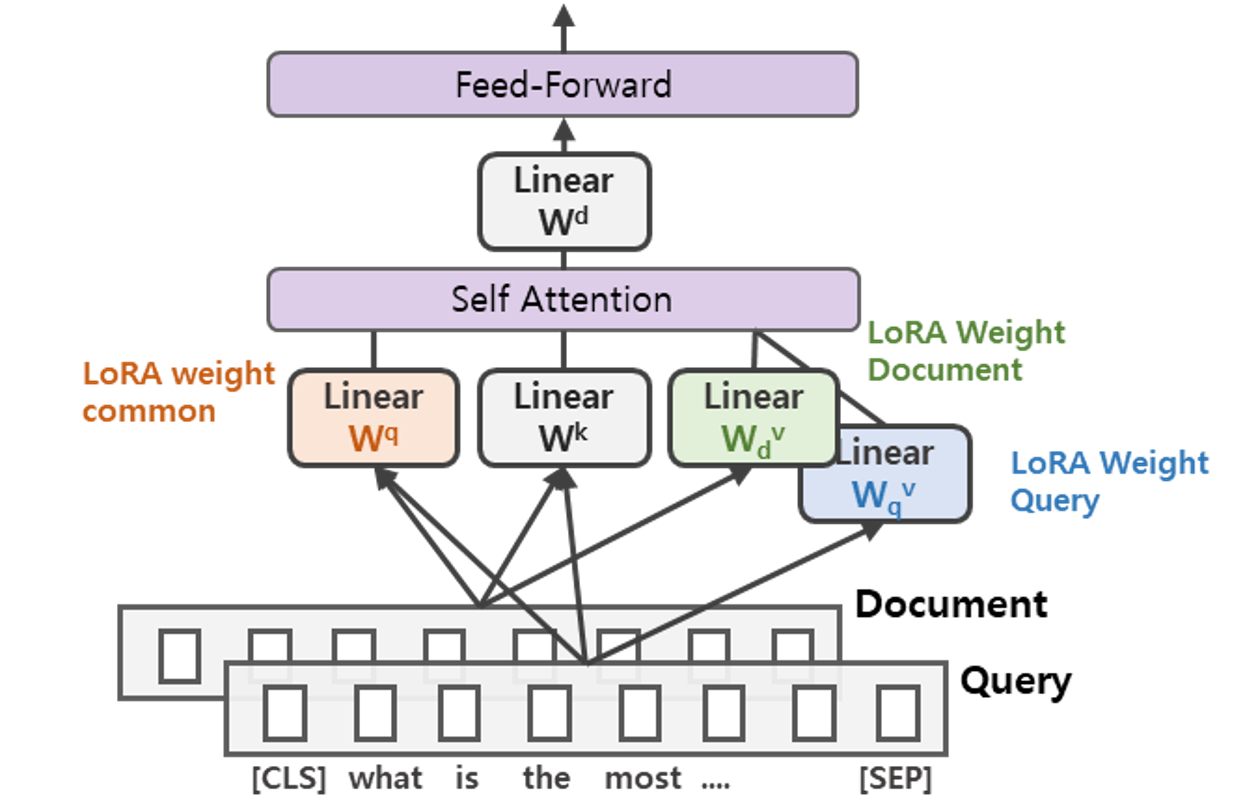}}
    \hfill
    \caption{The architectures of semi-Siamese prefix-tuning and semi-Siamese LoRA. (a) SS prefix-tuning utilizes both common prefixes and query/document specific prefixes. (b) SS LoRA utilizes a common query weight $W^{q}$ and query/document specific value weights $W^{v}_{q}$ and $W^{v}_{d}$.}
    \label{fig:semi_fig}
\end{figure*}

We propose three types of semi-Siamese LFT for bi-encoder NRMs.

\subsubsection{SS prefix-tuning}
To allow bi-encoder models to effectively process query-specific and document-specific information, we devised semi-Siamese prefix-tuning. As shown in Figure~{\ref{fig:semi_fig}}(a), we generate SS prefixes by summing up common prefixes with query-specific or document-specific prefixes:
\begin{align}
    h_{q,i} &= P_{\theta}[i,:] + P_{\theta_{q}}[i,:] \; if \; i \in P_{idx}  \label{eq: SS Prefix-tuning (average, query)} \\
    h_{d,i} &= P_{\theta}[i,:] + P_{\theta_{d}}[i,:] \; if \; i \in P_{idx}  \label{eq: SS Prefix-tuning (average, document)}
\end{align}
where $P_{\theta}[i,:]$ means the common prefix, $P_{\theta_{q}}[i,:]$ is the prefix for query, and $P_{\theta_{d}}[i,:]$ is the prefix for document. With this method, prefixes of query and document share the information through common prefixes while keeping their own characteristics in specific prefixes.
$P'_{\theta}$, the source input for MLP, is shared for $\theta$, $\theta_{q}$, and $\theta_{d}$.
We explored a few options for SS prefix-tuning. They are discussed in Appendix~\ref{Appendix:SSPrefix}, and the best performing option is presented here.

\subsubsection{SS LoRA}
We also devised semi-Siamese LoRA as illustrated in Figure~\ref{fig:semi_fig}(b). Because there are two types of LoRA weights, one for query and the other for value, we chose to use common LoRA weights for query projection and to use heterogeneous LoRA weights for value projection. In other words, we train $W^{v}_{q}$, $W^{v}_{d}$, and $W^{q}$ where $W^{v}_{q}$ and $W^{v}_{d}$ are self attention's value projection matrix for query and document respectively, and $W^{q}$ is the query projection matrix for both query and document \footnote{\textit{q} in the superscript represents the projection matrix type of self-attention. On the other hand, \textit{q} in the subscript represents the type of input to NRMs.}. 
We also explored a few options for SS LoRA. They are discussed in Appendix~\ref{Appendix:SSLoRA}, and the best performing option is presented here.


\subsubsection{SS sequential hybrid}
The sequential hybrid can be modified for learning semi-Siamese networks. Because we have two individual LFTs in the sequential hybrid method, we have three options for applying semi-Siamese: apply semi-Siamese to prefix-tuning only, LoRA only, or both. We have lightly investigated all three options without any tuning and have found that they achieved comparable performances. In our experiment section, we show the results for applying SS to LoRA only. Therefore, evaluation results for \textit{SS LoRA $\rightarrow$ Prefix-tuning} and \textit{Prefix-tuning $\rightarrow$ SS LoRA } are provided.

\section{Experiment}

 \begin{table*}[t]
	\caption{Evaluation results of lightweight fine-tuning: Cross-encoder. Note: \normalfont{$^{*} p \leq 0.05$, $^{**} p \leq 0.01$ (1-tailed).}}
	\label{tab:lwt cross-encoder}
	\adjustbox{max width=\linewidth}{%
	\begin{tabular}{ccc|ccc|ccc|ccc}
		\toprule
		\multirow{2}{*}{Model} & Fine-tuning & \# of trainable &
		\multicolumn{3}{c|}{\textbf{Robust04}} & \multicolumn{3}{c|}{\textbf{ClueWeb09b}} & \multicolumn{3}{c}{\textbf{MS-MARCO}} \\
		& method & parameters & P@20 & NDCG@5 & NDCG@20 & P@20 & NDCG@5 & NDCG@20 & P@20 & NDCG@5 & NDCG@20 \\
		\midrule
		\midrule
		& \textit{Full FT} & 110M & 0.3966 & 0.5310 & 0.4646 & 0.3059 &0.3101 & 0.2938 & 0.5725 & 0.6536& 0.5785 \\ \cline{2-12}
		& \textit{Prompt-tuning} & 8K & 0.3548 & 0.4733 & 0.4161 & 0.2917 & 0.2715 & 0.2715 & 0.5155 & 0.5075 & 0.4937 \\
		& \textit{Prefix-tuning} & 0.1M & 0.3936 & 0.5195 & 0.4602 & 0.3074  & 0.3040 & 0.2914 & 0.5531 & 0.5979 & 0.5435 \\	
		\cline{2-12}
		monoBERT & \textit{LoRA} & 0.6M & 0.3938 & 0.5296 & 0.4616 & 0.3150$^{*}$ & 0.3256$^{*}$ & 0.3054$^{**}$ & 0.5756 & 0.6656 & 0.5835 \\ 
		(cross) & \textit{LoRA$+$} & 0.9M & {\bf{0.4012}} & {\bf{0.5355}} & {\bf{0.4691}} & {\bf{0.3175}}$^{*}$ & 0.3154 & {0.3029}$^{*}$ & {\bf{0.5826}}$^{*}$ & {\bf{0.6662}} & {\bf{0.5887}}$^{*}$ \\
		\cline{2-12}
		 & \textit{Prefix-tuning $\rightarrow$ LoRA} & 0.7M & 0.3980 & 0.5241 & 0.4636 & 0.3046 & 0.3043 & 0.2911 & 0.5570 & 0.6055 & 0.5478 \\ 
		 & \textit{LoRA $\rightarrow$ Prefix-tuning} & 0.7M &  0.3960 & 0.5286 & 0.4628 & 0.3163$^{*}$ & \bf{0.3265}$^{*}$ & {\bf{0.3070}}$^{**}$ & {0.5748} & 0.6545 & 0.5809 \\ 
		\cline{2-12}
		 & \textit{Improvement (\%)} & - & 1.16\% & 0.85\% & 0.97\% & 3.79\% & 5.29\% & 4.49\% & 1.76\% & 1.93\% & 1.76\% \\
       \bottomrule
       
	\end{tabular}
   }
\end{table*}

\subsection{Experimental Setup}

\subsubsection{Datasets and metrics}
We conduct our experiments on Robust04~\cite{voorhees2004overview}, WebTrack 2009 (ClueWeb09b)~\cite{callan2009clueweb09}, and MS-MARCO~\cite{nguyen2016ms} datasets as in \cite{macavaney2019cedr}. Following \citet{huston2014parameters}, we divide each of Robust04 and ClueWeb09b into five folds and use three folds for training, one for validation, and the remaining one for test. For MS-MARCO, we used the pre-assigned datasets for training, validation, and test. We use document collections from TREC discs 4 and 5\footnote{520K documents, 7.5K triplet data samples, https://trec.nist.gov/data-disks.html} of Robust04 and from ClueWeb09b\footnote{50M web pages, 4.5K triplet data samples, https://lemurproject.org/clueweb09/} of WebTrack 2009. We also use MS-MARCO document collections.\footnote{22G documents, 372K triplet data samples, https://microsoft.github.io/msmarco/TREC-Deep-Learning-2019} For evaluation metrics, we used P@20, nDCG@5, and nDCG@20.

\subsubsection{Baseline models}
We implement three BERT-based NRMs, monoBERT \cite{nogueira2019passage}, ColBERT \cite{khattab2020colbert}, and TwinBERT \cite{lu2020twinbert}. 
monoBERT is a cross-encoder model, and the other two are bi-encoder models.
We consider the full fine-tuning performance of these models as the baseline performance.
We compare LFT and full FT by inspecting the improvements of LFT over full FT.

\subsubsection{Training and optimization}
When fine-tuning BERT-based models on down-stream tasks, the convention of training only three epochs was shown to be insufficient~\cite{zhang2020revisiting}. We set the maximum epoch as 30 for Robust04 and ClueWeb09b and as 10 for MS-MARCO. We select the checkpoint whose validation score is highest among the checkpoints of all epochs. For all experiments, we used an Adam~\cite{kingma2015adam} optimizer. For each method, we used appropriately selected hyper-parameters. For fine-tuning, we set the learning rate ($lr$) for the ranker as 1e-4 and $lr$ for BERT as 2e-5. For prefix-tuning, we used $lr$ of 1e-4 for prefix parameters and ranker weights. For LoRA, we used $lr$ of 1e-4 for both LoRA weights and ranker weights. The detailed setting of hyper-parameters are showed in Table \ref{tab: HP} in Appendix C. We repeated each experiment three times with three different random seeds. The results in the tables are calculated by averaging the test scores of all folds of the three experiments.
For statistical analysis, we performed one-tailed t-test under the assumption of homoscedasticity. We compared the three performance values of each LFT method with the three performance values of the corresponding full fine-tuning.

\subsubsection{Implementation}
Our experiments are implemented with Python 3 and PyTorch 1. We use a popular transformer library\footnote{https://github.com/huggingface/transformers} for the pre-trained BERT model. We used 10 RTX3090 GPUs each of which has 25.6G memory.

\subsection{LFT Results for Cross-encoder}
Table \ref{tab:lwt cross-encoder} shows the evaluation results for monoBERT. Here, we compare the performances of full fine-tuning and lightweight fine-tuning. We also show the number of fine-tuning parameters of each fine-tuning method. Prompt-tuning trains the least amount of parameters of 8K, and LoRA$+$ trains the largest amount of 0.9M. All LFT methods train less than 1M of parameters and thus less than 1\% of the BERT's 110M weight parameters.

\subsubsection{Prompt-tuning and Prefix-tuning}
Prompt-tuning simply appends the prefix to the input by training only 8K parameters, and the performance is inferior to the baseline of full fine-tuning by about 10\%. \citet{lester2021power} showed that prompt-tuning could achieve a comparable performance to full fine-tuning on SuperGLUE tasks, but we can observe that training only prompts is not sufficient for NRMs.
As for the prefix-tuning, it degrades the performance of full fine-tuning by about 3\% for MS-MARCO but it achieves comparable performance for the other two datasets. From the results, we can observe that prefix-tuning achieves comparable performance to full fine-tuning for Robust04 and ClueWeb09b that have short queries, but not for MS-MARCO.

\subsubsection{LoRA and LoRA$+$}
For cross-encoder, LoRA and LoRA$+$ perform significantly better than full fine-tuning for all three datasets with up to a 5\% of improvement. In particular, LoRA$+$ achieves the best performance for 7 out of 9 evaluation cases.
We can explain this result with two reasons. First, because LoRA methods freeze the pre-trained BERT and train only a small amount of the augmented weights, they act as a regularizer with a better generalization. Second, we can infer that the number of trainable parameters is not large enough for LoRA. As we can see in the Table \ref{tab:lwt cross-encoder}, the performance tends to increase as the number of parameters increases. LoRA has more trainable parameters than prefix-tuning and LoRA$+$ has 1.5 times more trainable parameters than LoRA. 

 \begin{table*}[t]
	\caption{Evaluation results of lightweight fine-tuning: Bi-encoders. Note: \normalfont{$^{*} p \leq 0.05$, $^{**} p \leq 0.01$ (1-tailed).}}
	\label{tab:lwt bi-encoder}
	\adjustbox{max width=\linewidth}{%
	\begin{tabular}{ccc|ccc|ccc|ccc}
		\toprule
		\multirow{2}{*}{Model} & Fine-tuning & \# of trainable &
		\multicolumn{3}{c|}{\textbf{Robust04}} & \multicolumn{3}{c|}{\textbf{ClueWeb09b}} & \multicolumn{3}{c}{\textbf{MS-MARCO}} \\
		& method & parameters & P@20 & NDCG@5 & NDCG@20 & P@20 & NDCG@5 & NDCG@20 & P@20 & NDCG@5 & NDCG@20 \\
		\midrule
		\midrule
		 & \textit{Full FT} & 110M & 0.3059 & \bf{0.3711} & 0.3468& 0.1846 & 0.1169 & 0.1471 & 0.5217 & \bf{0.5807} & 0.5141 \\ \cline{2-12}
		 & \textit{Prompt-tuning} & 8K & 0.2562 & 0.2666 & 0.2756 & 0.1900 & 0.1248 & 0.1558 & 0.4419 & 0.3659 & 0.3797 \\
		 & \textit{Prefix-tuning} & 0.1M & {\bf{0.3117}} & 0.3647 & {0.3496} & {0.2437}$^{**}$ & {0.2065}$^{**}$ & {0.2168}$^{**}$ & 0.5132 & 0.5630 & 0.5014 \\ 
		 \cline{2-12}
		 TwinBERT & \textit{LoRA} & 0.6M & 0.3090 & 0.3639 & 0.3484 &0.1851 & 0.1224 & 0.1530 & {0.5326} & 0.5458 & {0.5154} \\
		 (bi) & \textit{LoRA$+$} & 0.9M & 0.3056& 0.3656& 0.3451& {0.2028}$^{*}$ & {0.1740}$^{**}$ & {0.1814}$^{**}$ & {0.5318} & 0.5730 & {\bf{0.5207}} \\ 
		 \cline{2-12}
		 & \textit{Prefix-tuning $\rightarrow$ LoRA} & 0.7M & {0.3107} & 0.3667 & {0.3484} & {\bf{0.2447}}$^{**}$ & {\bf{0.2155}}$^{**}$ & {\bf{0.2192}}$^{**}$ & 0.5221 & 0.5485 & 0.5048 \\ 
		 & \textit{LoRA $\rightarrow$ Prefix-tuning} & 0.7M & {0.3102} & 0.3685 & {\bf{0.3508}} & 0.1964 & 0.1402$^{**}$ & 0.1647$^{*}$ & {\bf{0.5334}} & 0.5515 & {0.5184} \\ 
		 \cline{2-12}
		 & \textit{Improvement (\%)} & - & 1.90\% & - & 1.15\% & 32.56\% & 84.35\% & 49.01\% & 2.24\% & - & 1.28\% \\
		\midrule
		\midrule
		 & \textit{Full FT} & 110M & 0.3335 & 0.3990 & 0.3760 & 0.2659 & 0.2507 & 0.2541 & 0.5566 &0.6121 & 0.5484 \\ \cline{2-12}
		 & \textit{Prompt-tuning} & 8K & 0.3077 & 0.3446 & 0.3404 & 0.2440 & 0.1965 & 0.2177 & 0.5275 & 0.4942 & 0.4965 \\
		 & \textit{Prefix-tuning} & 0.1M & {\bf{0.3429}}$^{*}$ &  {0.4084} & {\bf{0.3865}}$^{**}$ & {\bf{0.2695}} & {\bf{0.2571}} & {\bf{0.2544}} & 0.5577 & 0.6221 & 0.5556 \\ 
		 \cline{2-12}
		 ColBERT & \textit{LoRA} & 0.6M & {0.3386} & 0.4021 & {0.3818} & 0.2644 & 0.2373 & 0.2498 & {0.5601} & {\bf{0.6360}}$^{*}$ & {0.5566} \\
		 (bi) & \textit{LoRA$+$} & 0.9M & 0.3385 & 0.3997 & 0.3804 & 0.2606 & 0.2453 &0.2481 & 0.5574 & {0.6292}$^{*}$ & 0.5543 \\
		 \cline{2-12}
		 & \textit{Prefix-tuning $\rightarrow$ LoRA} & 0.7M & {0.3411} & {\bf{0.4103}} & {0.3855}$^{*}$ & {0.2675} & {0.2511} & 0.2512 & {0.5605} & 0.6200 & {0.5580} \\ 
		 & \textit{LoRA $\rightarrow$ Prefix-tuning} & 0.7M & 0.3360 & {0.4025} & 0.3803 & 0.2639 & 0.2436 & 0.2525 & {\bf{0.5620}} & {0.6331} & {\bf{0.5595}}$^{*}$ \\ 
		 \cline{2-12}
		 & \textit{Improvement (\%)} & - & 2.82\% & 2.83\% & 2.79\% & 1.35\% & 2.55\% & 0.12\% & 0.97\% & 3.90\% & 2.02\% \\
       \bottomrule
	\end{tabular}
    }
\end{table*}

 \begin{table*}[t]
	\caption{Evaluation results of semi-Siamese (SS) lightweight fine-tuning (LFT) for Bi-encoders - semi-Siamese provides positive improvements for most of the evaluation cases of Robust04 and ClueWeb09b datasets whose queries are short. The gain is very large for ClueWeb09b whose queries are the shortest among the three datasets. Note: \normalfont{$^{*} p \leq 0.05$, $^{**} p \leq 0.01$ (1-tailed).}}
	\label{tab:SS lwt}
	\adjustbox{max width=\linewidth}{%
	\begin{tabular}{ccc|ccc|ccc|ccc}
		\toprule
		\multirow{2}{*}{Model} & Fine-tuning & \# of trainable &
		\multicolumn{3}{c|}{\textbf{Robust04}} & \multicolumn{3}{c|}{\textbf{ClueWeb09b}} & \multicolumn{3}{c}{\textbf{MS-MARCO}} \\
		& method & parameters & P@20 & NDCG@5 & NDCG@20 & P@20 & NDCG@5 & NDCG@20 & P@20 & NDCG@5 & NDCG@20 \\
		\midrule
		\midrule
		 & \textit{Full FT} & 110M & 0.3059 & 0.3711 & 0.3468& 0.1846 & 0.1169 & 0.1471 & 0.5217 & \bf{0.5807} & 0.5141 \\ 
		 \cline{2-12}
		 & \textit{LFT (best)} & $\le$ 0.9M & 0.3117 & 0.3685 & 0.3508 & {0.2447}$^{**}$ & {0.2155}$^{**}$ & {0.2192}$^{**}$ & {\bf{0.5334}} & 0.5730 & {\bf{0.5207}} \\ 
		 \cline{2-12}
		 TwinBERT & \textit{SS Prefix-tuning} & 0.1M & 0.3109 & 0.3678 & 0.3492 & 0.2223$^{**}$ & 0.1721$^{**}$ & 0.1942$^{**}$ & 0.5155 & 0.5229 & 0.4918 \\
		 (bi) & \textit{SS LoRA} & 0.9M & {0.3117} & {0.3713} & {0.3511} & 0.1926 & 0.1341$^{*}$ & 0.1603 & 0.5240 & 0.5778 & 0.5129 \\
		 & \textit{Prefix-tuning $\rightarrow$ SS LoRA} & 1M & 0.3104 & 0.3637 & 0.3473 & {\bf{0.2464}}$^{**}$ & {\bf{0.2170}}$^{**}$ & {\bf{0.2197}}$^{**}$ & 0.5140 & 0.5419 & 0.4983 \\
		 & \textit{SS LoRA $\rightarrow$ Prefix-tuning} & 1M & {\bf{0.3146}} & {\bf{0.3809}} & {\bf{0.3575}}$^{*}$ & 0.2012$^{*}$ & 0.1465$^{**}$ & 0.1689$^{*}$ & {0.5252} & 0.5685 & {0.5142} \\
		 \cline{2-12}
		 & \textit{Improvement (\%)} &  & 2.84\% & 2.64\% & 3.09\% & 33.48\% & 85.63\% & 49.35\% & 2.24\% & - & 1.28\% \\		 
		 
		\midrule
		\midrule
		 & \textit{Full FT} & 110M & 0.3335 & 0.3990 & 0.3760 & 0.2659 & 0.2507& 0.2541 & 0.5566 & 0.6121 & 0.5484 \\ 
		 \cline{2-12}
		 & \textit{LFT (best)} & $\le$ 0.9M  & {0.3429}$^{*}$ & {0.4103} & {0.3865}$^{**}$ & {0.2695} & {0.2571} & {0.2544} & {0.5620} & {\bf{0.6360}}$^{*}$ & {0.5595}$^{*}$ \\ 
		 \cline{2-12}
		 ColBERT & \textit{SS Prefix-tuning} & 0.1M & {\bf{0.3442}}$^{*}$ & {\bf{0.4113}} & {\bf{0.3883}}$^{**}$ & {\bf{0.2950}}$^{**}$ & {\bf{0.2914}}$^{*}$ & {\bf{0.2835}}$^{**}$ & 0.5554 & 0.6125
		 & 0.5526 \\
		 (bi) & \textit{SS LoRA} & 0.9M & 0.3406$^{*}$ & 0.4042 & 0.3823 & 0.2615 & 0.2470 & 0.2500 & {\bf{0.5647}} & {0.6289} & 0.5587 \\
		 & \textit{Prefix-tuning $\rightarrow$ SS LoRA} & 1M & 0.3417 & 0.4083 & 0.3850$^{*}$ & 0.2679 & 0.2512 & 0.2514 & 0.5593 & 0.6251 & {\bf{0.5596}} \\
		 & \textit{SS LoRA $\rightarrow$ Prefix-tuning} & 1M & 0.3420 & 0.4094 & 0.3845 & 0.2642 & 0.2474 & 0.2520 & 0.5577 & 0.6173 & 0.5545 \\
		 \cline{2-12}
		 & \textit{Improvement (\%)} &  & 3.21\% & 3.08\% & 3.27\% & 10.94\% & 16.23\% & 11.57\% & 1.46\% & 3.90\% & 2.04\% \\
       \bottomrule
	\end{tabular}
    }
\end{table*}

\subsubsection{Hybrid LFT}
Prefix-tuning and LoRA can be used together because they train parameters in representation dimension and weight dimension, respectively. We combined the two LFT methods by adopting a sequential augmentation of BERT. Overall, hybrid LFT methods show better performance than prefix-tuning, but do not outperform LoRA$+$.

\subsection{LFT Results for Bi-encoders}
Table \ref{tab:lwt bi-encoder} shows the evaluation results of LFT for bi-encoders, TwinBERT and ColBERT. The effectiveness of LFT methods exhibits a quite different pattern compared to the case of cross-encoder, and the results and their explanations are provided below. 

\subsubsection{Prompt-tuning and Prefix-tuning}
As in the cross-encoder, prompt-tuning shows degenerate results compared to the full fine-tuning baseline. The only exception is the case of TwinBERT on ClueWeb09b, but perhaps it is due to the poor baseline performance. Therefore, we confirm that prompt-tuning is not adequate for the document ranking tasks and we exclude it from the Semi-Siamese experiments in the next section.

In contrast to the cross-encoder results, prefix-tuning achieves significant improvements over the full fine-tuning for the datasets of Robust04 and ClueWeb09b that have short queries. It also achieves a comparable performance to full fine-tuning for MS-MARCO.
We first focus on the results of Robust04 and ClueWeb09b that consist of short and keyword-based queries. Prefix-tuning shows the best performance for most of the cases with improvements over full fine-tuning by large margins of up to 76.65\%. Prefix-tuning also outperforms LoRA and LoRA$+$. We attribute this behavior to the way of bi-encoder's encoding and the characteristics of the datasets. As shown in Table \ref{tab:query_length}, queries in Robust04 and ClueWeb09b are relatively short and keyword-based. Since bi-encoder models encode query and document separately, the model needs to extract the contextual information from either a query or a document. For query, however, it can be very short and thus BERT can fail to extract any meaningful contextual information. In this case, prefix-tuning has the advantage of adding task-specific and meaningful contextual information into the representation embeddings. For the dataset of MS-MARCO, however, queries are relatively long and the benefit of prefix-tuning becomes smaller. We discuss this result further in Section 5. 

\subsubsection{LoRA and LoRA$+$}
Unlike in the case of cross-encoder, LoRA and LoRA$+$ do not dominantly outperform the other methods. LoRA and LoRA$+$ show a varying performance according to the characteristics of dataset.
In the result of the cross-encoder, LoRA$+$ shows the best performance for Robust04 and ClueWeb09b. For bi-encoder, however, the performance of LoRA and LoRA$+$ is relatively inferior when compared to prefix-tuning.
For MS-MARCO that has long queries, LoRA or LoRA$+$ can perform better than full fine-tuning and prefix-tuning.
Interestingly, LoRA$+$ performs better than LoRA for cross-encoder, but LoRA performs better than LoRA$+$ for bi-encoders.
LoRA$+$ uses 50\% more weight than LoRA for learning, indicating that this difference worked positively for cross-encoder and negatively for bi-encoders.
Since cross-encoder models do self-attention between query and document, learning is likely to be more complex than in the bi-encoder models that do not use self-attention between query and document.
Therefore, it can be assumed that it is advantageous to use more parameters for the cross-encoder models.


\subsubsection{Hybrid LFT}
Since prefix-tuning shows outstanding performance, and LoRA also performs better than full fine-tuning, the combination of two LFT methods might lead to an additional performance improvement. In Table \ref{tab:lwt bi-encoder}, we included the performance of hybrid LFT methods. The results show the possibility of hybrid methods to improve the performance of single LFT methods.

\subsection{Semi-Siamese LFT Results for Bi-encoders}

LFT methods easily outperform full fine-tuning on document ranking. We further improve LFT methods by applying semi-Siamese networks. SS LFT can handle query and document slightly differently to reflect their distinct characteristics while maintaining the original BERT without any modification to enable consistent encoding over query and document. Table \ref{tab:SS lwt} shows that SS LFT methods can improve the best performing LFT or full fine-tuning in most cases.

\subsubsection{SS Prefix-tuning}
SS prefix-tuning performs significantly better than the other methods when adopted for ColBERT on Robust04 and ClueWeb09b, improving LFT's best performance from 3.08\% to 16.23\%. Also in other cases, SS prefix-tuning shows the possibility of improving prefix-tuning. We can say that semi-Siamese networks help bi-encoder models effectively process information in query and document.

\subsubsection{SS LoRA}
As shown in the table \ref{tab:SS lwt}, SS LoRA outperforms LoRA in most cases, indicating that semi-Siamese networks allow LoRA for bi-encoder models to better estimate relevance scores. We infer that using different LoRA weights for the query and document representations induces a performance improvement by giving query and document more capacity to focus on query-specific or document-specific information.

\subsubsection{SS Hybrid LFT}
We have shown that SS LoRA and SS Prefix-tuning can improve the document ranking performance. We have also shown that prefix-tuning and LoRA can complement each other, increasing the performance when used together. Therefore, we combined prefix-tuning and SS LoRA by sequentially training parameters of each method. From the Table \ref{tab:SS lwt}, we can see that our hybrid methods perform best in many cases bringing up to 85.63\% of improvement over full fine-tuning. SS could have been applied to prefix-tuning as well, but we have found that SS LoRA is sufficient for hybrid LFT.

\section{Discussion}

\subsection{Cross-encoder vs. Bi-encoder}
\begin{figure}
    \centering
    \includegraphics[width=0.85\linewidth]{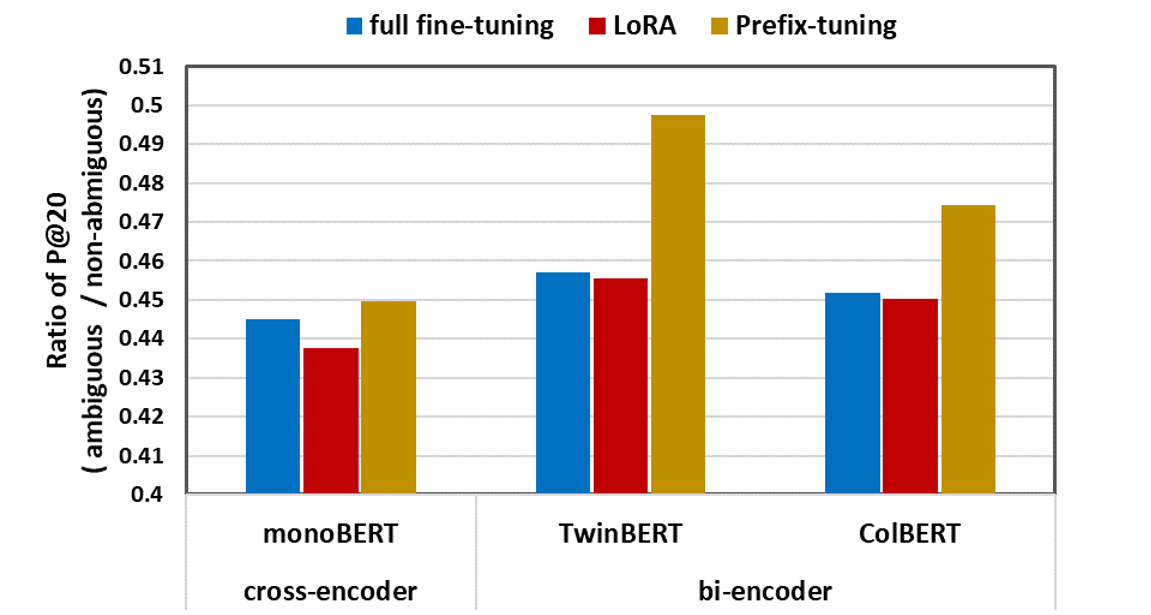}
    \caption{For ClueWeb09b, the P@20 performance ratio between ambiguous queries and non-ambiguous queries is shown. Semi-Siamese was not applied. For bi-encoders, prefix-tuning clearly shows relatively better performance for ambiguous queries.}
    \label{fig:ambiguous_query}
\end{figure}

For the cross-encoder results shown in Table~\ref{tab:lwt cross-encoder}, clearly LoRA based models outperform prefix-tuning based models. In fact, LoRA$+$ is the dominant LFT method that achieves the best performance for 7 out of 9 evaluations. For the bi-encoder results shown in Table~\ref{tab:lwt bi-encoder} and Table~\ref{tab:SS lwt}, however, we can observe that prefix-tuning based models clearly outperform LoRA based models at least for Robust04 and ClueWeb09b. In fact, SS prefix-tuning is the best performing model for 6 out of 6 cases for ColBERT in Table~\ref{tab:SS lwt}. 

We provide a possible explanation for the phenomenon. First, we can consider the cross-encoder case. For cross-encoder, query and document are used together as a single input to the BERT. Because the pair is used together, the direct relationship or contextual information between the query and the document can be evaluated by the language model. In this case, the value of prepending with task-specific prefixes can become insignificant, and probably it suffices to fine-tune the weight-related parameters such as LoRA weights. Second, we can consider the bi-encoder case. For Robust04 and ClueWeb09b, queries are short. Therefore, bi-encoder might not be able to create effective contextual representations for the query part. Note that no document is visible to BERT when processing a query. Then, it can be more important to provide task-specific information by processing the entire training dataset, such that at least task information can be used for creating contextual query representation that is effective. This can be achieved with the prefixes. 

As a further analysis, we have analyzed ClueWeb09b dataset. We divided one-word queries of ClueWeb09b, that has the shortest queries, into ambiguous queries and non-ambiguous queries using pywsd library\footnote{https://pypi.org/project/pywsd/}. We only included one-word queries because a high-confidence ambiguity classification is possible for one-word queries. The results are shown in Figure~\ref{fig:ambiguous_query}. We can see that prefix-tuning is effective for improving the performance of ambiguous queries while full FT and LoRA are not effective.


\subsection{Hybrid: Concurrent Learning vs. Sequential Learning}
\begin{figure}
    \centering
    \includegraphics[width=0.83\linewidth]{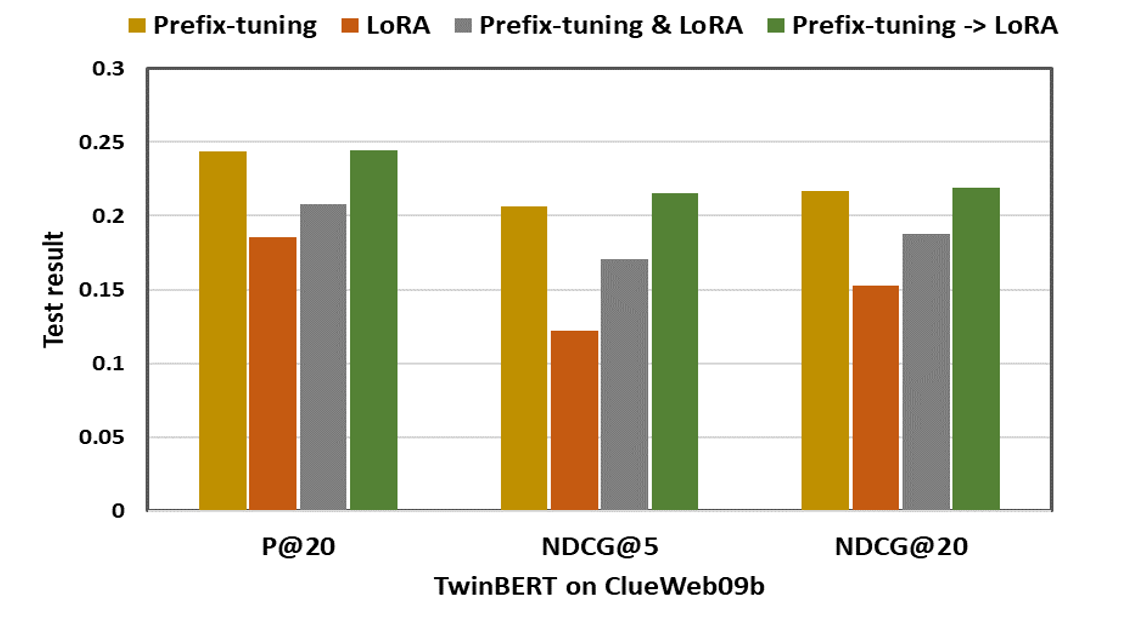}
    \caption{Comparison of prefix-tuning, LoRA, pefix-tuning \& LoRA (concurrent learning), and prefix-tuning $\rightarrow$ LoRA (sequential learning). It can be seen that concurrent learning can cause a performance degradation when there is a large performance gap between the two individual methods. }
    \label{fig:sequential_explain}
\end{figure}

When combining two LFT methods to create a hybrid method, we have chosen to sequentially train the two LFT modules. The other option is to concurrently train the two LFT modules. The reason for choosing the sequential training can be explained with Figure~\ref{fig:sequential_explain}. In the example, prefix-tuning performs well and LoRA does not perform well. If we choose concurrent learning, we end up with a performance that is worse that prefix-tuning only. This indicates that concurrent learning can hinder a proper learning of individual LFTs, especially when the performance gap between the two individual methods is large. To avoid this problem, we have chosen to train them in a sequential way. By considering the two possible orders of sequential training, we are likely to preserve the larger gain of the two individual gains. We didn't necessarily obtain improved performance for all the cases by adopting sequential hybrid learning, but quite often we were able to obtain improvements over the individual approaches in the case of bi-encoders.




\section{conclusion}
We have shown the effectiveness of adopting lightweight fine-tuning methods such as prefix-tuning and LoRA to replace the full fine-tuning of the existing bi-encoder NRMs. We have also shown how semi-Siamese networks can be used to achieve a significant performance improvement when the queries are very short. 
Our semi-Siamese architecture is also efficient in terms of storage and memory requirements thanks to the use of lightweight fine-tuning for creating two slightly different networks.

\begin{acks}
This work was supported by Naver corporation (Development of an Improved Neural Ranking Model.), National Research Foundation of Korea~(NRF) grant funded by the Korea government~(MSIT) (No. NRF-2020R1A2C2007139), and IITP grant funded by the Korea government~(No. 2021-0-01343, Artificial Intelligence Graduate School Program~(Seoul National University)).
\end{acks}

\bibliographystyle{ACM-Reference-Format}
\bibliography{acmart}

\appendix
\section{Variants of SS Prefix-tuning}
\label{Appendix:SSPrefix}

\begin{table*}[t]
	\caption{Variants of SS prefix-tuning. We compare performances of prefix-tuning and three SS prefix-tuning methods. Among SS prefix-tuning methods, the method that shares information between query and document by averaging common prefixes is our suggested method and it is used for obtaining the results in Table~\ref{tab:SS lwt}.}
	\label{tab: Variants of SS Prefix-tuning}
	\adjustbox{max width=\linewidth}{%
	\begin{tabular}{cc|ccc|ccc|ccc}
		\toprule
		\multirow{2}{*}{Model} & Prefix sharing &
		\multicolumn{3}{c|}{\textbf{Robust04}} & \multicolumn{3}{c|}{\textbf{Clueweb09b}} &
		\multicolumn{3}{c}{\textbf{MS-MARCO}} \\
		& (query-document) & P@20 & NDCG@5 & NDCG@20 & P@20 & NDCG@5 & NDCG@20 & P@20 & NDCG@5 & NDCG@20 \\
		\midrule
		\midrule
		 & \textit{All} & 0.3117 & 0.3647 & 0.3496 & 0.2437 & 0.2065 & 0.2168 & 0.5132 & 0.5630 & 0.5014 \\ 
		 \cline{2-11}
		 TwinBERT & \textit{Average*} & 0.3109 & 0.3678 & 0.3492 & 0.2223 & 0.1721 & 0.1942 & 0.5155 & 0.5229 & 0.4918 \\ 
		 (bi) & \textit{Concatenation} & 0.3101 & 0.3685 & 0.3505 & 0.2105 & 0.1638 & 0.1836 & 0.5174 & 0.5347 & 0.4970 \\ 
		 & \textit{None} & 0.3085 & 0.3597 & 0.3453 & 0.2036 & 0.1554 & 0.1746 & 0.5015 & 0.4955 & 0.4716 \\ 
		 		 
		\midrule
		\midrule
		 & \textit{All} & 0.3429 & 0.4084 & 0.3865 & 0.2695 & 0.2571 & 0.2544 & 0.5577 & 0.6221 & 0.5556 \\ 
		 \cline{2-11}
		 ColBERT & \textit{Average*}  & 0.3442 & 0.3472 & 0.3459 & 0.2950 & 0.2914 & 0.2835 & 0.5554 & 0.6125 & 0.5526\\ 
		 (bi) & \textit{Concatenation} & 0.3373 & 0.4071 & 0.3813 & 0.2807 & 0.2755 & 0.2694 & 0.5605 & 0.6221 & 0.5588 \\
		 & \textit{None} & 0.3384 & 0.4057 & 0.3823 & 0.2774 & 0.2682 & 0.2663 & 0.5546 & 0.5920 & 0.5392 \\ 
       \bottomrule
	\end{tabular}
    }
\end{table*}

We experimented with three SS Prefix-tuning methods and compared them with Siamese Prefix-tuning.
As the first method, we generated prefixes by summing up common prefixes with query-specific or document-specific prefixes as in the Equations {\ref{eq: SS Prefix-tuning (average, query)}} and {\ref{eq: SS Prefix-tuning (average, document)}}. This method is \textit{SS Prefix-tuning} in Table {\ref{tab:SS lwt}}.
As the second method, we generated $k$ common prefixes for query and document and concatenate them with query-specific or document-specific prefixes as below equations where $\theta_{q}$ and $\theta_{d}$ indicate parameters for query and document respectively, $\theta$ is the parameter common to both query and document, and $k$ stands for the number of query-specific and document-specific prefixes.
\begin{equation}
\label{eq: SS Prefix-tuning (query)}
\begin{aligned}
    h_{q,i} &= \begin{cases}
            P_{\theta_{q}}[i,:] \; if \; i \leq k \; and \; i \in P_{idx} \\
            P_{\theta}[i,:] \; if \; i > k \; and \; i \in P_{idx}
                \end{cases} \\
\end{aligned}
\end{equation}
\begin{equation}
\label{eq: SS Prefix-tuning (document)}
\begin{aligned}
    h_{d,i} &= \begin{cases}
            P_{\theta_{d}}[i,:] \; if \; i \leq k \; and \; i \in P_{idx} \\
            P_{\theta}[i,:] \; if \; i > k \; and \; i \in P_{idx}
                \end{cases} 
\end{aligned}
\end{equation}
As the third method, we generated prefixes that did not share information between query and document as in Equations \ref{eq: SS Prefix-tuning (hetero, query)} and \ref{eq: SS Prefix-tuning (hetero, document)}. In this case, prefixes for query and document are generated using different weight parameters while sharing the source vectors.
\begin{align}
    h_{q,i} &= P_{\theta_{q}}[i,:] \; if \; i \in P_{idx} \label{eq: SS Prefix-tuning (hetero, query)} \\
    h_{d,i} &= P_{\theta_{d}}[i,:] \; if \; i \in P_{idx} \label{eq: SS Prefix-tuning (hetero, document)}
\end{align}
In Table~{\ref{tab: Variants of SS Prefix-tuning}}, we compare the performance of prefix-tuning variants. From this result, we confirm that the semi-Siamese methods perform generally well when the information is appropriately shared between query and document.

\section{Variants of SS LoRA}
\label{Appendix:SSLoRA}

\begin{table*}[t]
	\caption{Variants of SS LoRA. We compare performances of LoRA and three SS LoRA methods. Among the SS LoRA methods, the method that shares LoRA weights for the value projection with the asteroid mark is our suggested method and used for obtaining the results in Table~\ref{tab:SS lwt}.}
	\label{tab: Variants of SS LoRA}
	\adjustbox{max width=\linewidth}{%
	\begin{tabular}{cc|ccc|ccc|ccc}
		\toprule
		\multirow{2}{*}{Model} & LoRA weight sharing &
		\multicolumn{3}{c|}{\textbf{Robust04}} & \multicolumn{3}{c|}{\textbf{Clueweb09b}} &
		\multicolumn{3}{c}{\textbf{MS-MARCO}} \\
		& (query-document) & P@20 & NDCG@5 & NDCG@20 & P@20 & NDCG@5 & NDCG@20 & P@20 & NDCG@5 & NDCG@20 \\
		\midrule
		\midrule
		 & \textit{All} & 0.3090 & 0.3639 & 0.3484 & 0.1851 & 0.1224 & 0.1530 & 0.5326 & 0.5458 & 0.5154 \\ 
		 \cline{2-11}
		 TwinBERT & \textit{Value (hetero query)} & 0.3099 & 0.3666 & 0.3496 & 0.1882 & 0.1286 & 0.1543 & 0.5294 & 0.5816 & 0.5241 \\ 
		 (bi) & \textit{Query (hetero value)*} & 0.3117 & 0.3713 & 0.3511 & 0.1926 & 0.1341 & 0.1603 & 0.5240 & 0.5778 & 0.5129 \\ 
		 & \textit{None} & 0.3116 & 0.3790 & 0.3535 & 0.1870 & 0.1309 & 0.1553 & 0.5310 & 0.5630 & 0.5127 \\ 
		 		 
		\midrule
		\midrule
		 & \textit{All} & 0.3386 & 0.4021 & 0.3818 & 0.2644 & 0.2373 & 0.2498 & 0.5601 & 0.6360 & 0.5566 \\ 
		 \cline{2-11}
		 ColBERT & \textit{Value (hetero query)} & 0.3398 & 0.4071 & 0.3840 & 0.2611 & 0.2319 & 0.2455 & 0.5655 & 0.6263 & 0.5602 \\ 
		 (bi) & \textit{Query (hetero value)*} & 0.3406 & 0.4042 & 0.3823 & 0.2615 & 0.2470 & 0.2500 & 0.5647 & 0.6289 & 0.5587 \\ 
		 & \textit{None} & 0.3404 & 0.3947 & 0.3791 & 0.2595 & 0.2347 & 0.2435 & 0.5566 & 0.5998 & 0.5461 \\ 
       \bottomrule
	\end{tabular}
    }
\end{table*}

Because LoRA trains the LoRA weights for query and value, $W^{q}$ and $W^{v}$, we devised three SS LoRA methods.
As the first method, we used heterogeneous LoRA query weights for both query and document while using the same LoRA value weights for query and document. In other words, we trained $W^{q}_{q}$, $W^{q}_{d}$, and $W^{v}$. This method is written as \textit{SS LoRA (hetero query)} in Table \ref{tab: Variants of SS LoRA}. As the second method, we tried a different method where trained common LoRA query weights for query and document while training heterogeneous LoRA value weights for query and document. Here, we trained $W^{q}$, $W^{v}_{q}$, and $W^{v}_{d}$. This method is written as \textit{SS LoRA (hetero value)} in Table \ref{tab: Variants of SS LoRA}. As the third method, we trained different LoRA weights for query and document, letting $W^{q}_{q}$, $W^{q}_{d}$, $W^{v}_{q}$, and $W^{v}_{d}$ be trained. From the results in Table \ref{tab: Variants of SS LoRA}, we can observe that SS LoRA variants generally perform well. We infer that sharing information between query and document through LoRA weights is important for a better performance. When comparing SS LoRA variants, the method that shares LoRA query weights performs slightly better. We can infer that learning query-specific or document-specific value weights are more important in estimating relevance rather than using different query weights.

\section{Hyper-parameter setting}
\label{Appendix:hyperparameter}
We have lightly tuned the hyper-parameters of each method, and the values are shown in Table \ref{tab: HP}.
Additional explanations on LoRA alpha and LoRA dropout can be found in \cite{hu2021lora}.

\begin{table*}[b]
	\caption{Hyper-parameter settings of each dataset and fine-tuning methods.}
	\label{tab: HP}
	\adjustbox{max width=\linewidth}{%
	\begin{tabular}{ccccc}
		\toprule
		Method & Hyper-parameter & {\textbf{MS-MARCO}} & {\textbf{Robust04}} & {\textbf{Clueweb09b}} \\
		\midrule
		\midrule
		 & Optimizer & & Adam & \\ 
		 Common & Ranker $lr$ & & 0.0001 & \\ 
		 HP & Max epoch & 10 & 30 & 30 \\ 
		 & Batch size & & 16 & \\
		 & Weight decay & & 0 & \\
		 		 
		\midrule
		\midrule
		Full Fine-tuning & BERT $lr$ & & 0.00001 & \\
		
		\midrule
		\midrule
		 & Prefix length & & 10 & \\ 
		 Prefix- & Source dimension & & 256 & \\ 
		 tuning & Prefix $lr$ & & 0.0001 & \\ 
		 & Common prefix length  & & \multirow{2}{*}{5} & \\
		 & (SS Prefix-tuning (concat)) & & & \\
		 
		\midrule
		\midrule
		 & LoRA rank & & 16 & \\ 
		 LoRA & LoRA alpha & & 32 & \\ 
		 & LoRA dropout & & 0.1 & \\ 
		 & LoRA $lr$ & & 0.0001 & \\
      \bottomrule
	\end{tabular}
    }
\end{table*}

\end{document}